\documentclass[11pt,a4paper]{article}
\usepackage[hyperref]{acl2020}
\usepackage{times}
\usepackage{latexsym}
\usepackage{color}
\usepackage{multirow}
\usepackage{forest}
\usepackage{tikz}
\usepackage{algorithm}
\usepackage{algpseudocode}
\usepackage{amsmath}
\usepackage{graphics}
\usepackage{graphicx}
\usepackage{subfigure}
\usepackage{mwe}
\usepackage{amssymb}
\usepackage{bm}
\usepackage{amsmath}
\usepackage{makecell}
\usepackage{tabularx}
\usepackage{filecontents}
\usepackage{url}
\usepackage{pgfplots}
\usepgfplotslibrary{colorbrewer}
\usepackage{CJKutf8}
\pgfplotsset{width=10cm,compat=1.9}

\usepackage{microtype}

\aclfinalcopy
\def\aclpaperid{598}

\definecolor{electricviolet}{rgb}{0.56, 0.0, 1.0}
\definecolor{bblue}{HTML}{4F81BD}
\definecolor{rred}{HTML}{c4260b}
\definecolor{ggreen}{HTML}{098c1f}
\definecolor{ppurple}{HTML}{9F4C7C}
\definecolor{oorange}{HTML}{F79646}

\usetikzlibrary{arrows.meta,calc,decorations.markings,math,arrows.meta,patterns}
\usetikzlibrary{decorations.pathreplacing}
\usetikzlibrary{matrix}

\title{Norm-Based Curriculum Learning for Neural Machine Translation}
  
 \author{Xuebo Liu$^1$\thanks{~~Equal Contribution}~~~Houtim Lai$^{2}\footnotemark[1]$~~~Derek F. Wong$^1$\thanks{~~Corresponding author}~~~Lidia S. Chao$^1$\\
 $^1$NLP$^2$CT Lab, Department of Computer and Information Science, University of Macau \\
 $^2$NewTranx Information Technology, Shenzhen, China
  \\
  {\tt nlp2ct.xuebo@gmail.com, haotian.li@newtranx.com,} \\
  {\tt \{derekfw,lidiasc\}@um.edu.mo}\\}

\date{}

\begin{document}
\maketitle
\begin{abstract}
A neural machine translation (NMT) system is expensive to train, especially with high-resource settings.
As the NMT architectures become deeper and wider, this issue gets worse and worse.
In this paper, we aim to improve the efficiency of training an NMT by introducing a novel \textit{norm-based curriculum learning} method.
We use the norm (aka length or module) of a word embedding as a measure of 1) the difficulty of the sentence, 2) the competence of the model, and 3) the weight of the sentence.
The norm-based sentence difficulty takes the advantages of both linguistically motivated and model-based sentence difficulties.
It is easy to determine and contains learning-dependent features.
The norm-based model competence makes NMT learn the curriculum in a fully automated way, while the norm-based sentence weight further enhances the learning of the vector representation of the NMT. 
Experimental results for the WMT'14 English--German and WMT'17 Chinese--English translation tasks demonstrate that the proposed method outperforms strong baselines in terms of BLEU score (+1.17/+1.56) and training speedup (2.22x/3.33x).
\end{abstract}

\section{Introduction}
The past several years have witnessed the rapid development of neural machine translation (NMT) based on an encoder--decoder framework to translate natural languages \cite{kalchbrenner2013recurrent,sutskever2014sequence,bahdanau2014neural}.
Since NMT benefits from a massive amount of training data and works in a cross-lingual setting, it becomes much hungrier for training time than other natural language processing (NLP) tasks.

Based on self-attention networks \cite{Parikh:2016tz,Lin:2017ti}, Transformer \cite{DBLP:journals/corr/VaswaniSPUJGKP17} has become the most widely used architecture for NMT.
Recent studies on improving Transformer, e.g. deep models equipped with up to 30-layer encoders~\cite{Bapna:2018va,wu-etal-2019-depth,wang-etal-2019-learning-deep,zhang-etal-2019-improving}, and scaling NMTs which use a huge batch size to train with 128 GPUs~\cite{ott2018scaling,edunov-etal-2018-understanding}, 
face a challenge to the efficiency of their training.
Curriculum learning (CL), which aims to train machine learning models \emph{better} and \emph{faster}~\citep{bengio2009curriculum}, is gaining an intuitive appeal to both academic and industrial NMT systems.

The basic idea of CL is to train a model using examples ranging from ``easy'' to ``difficult'' in different learning stages, and thus the criterion of difficulty is vital to the selection of examples.
\citet{Zhang:2018vf} summarize two kinds of difficulty criteria in CL for NMT: 1) linguistically motivated sentence difficulty, e.g. sentence length, word frequency, and the number of coordinating conjunctions, which is easier to obtain~\citep{Kocmi:2017tw,cbcl}; 2) model-based sentence difficulty, e.g. sentence uncertainties derived from independent language models or the models trained in previous time steps or epochs, which tends to be intuitively effective but costly~\citep{Zhang:2016ue,Kumar:2019vn,Zhang:2019wg,UACL20}.

In this paper, we propose a novel norm-based criterion for the difficulty of a sentence, which takes advantage of both model-based and linguistically motivated difficulty features.
We observe that the norms of the word vectors trained on simple neural networks are expressive enough to model the two features, which are easy to obtain while possessing learning-dependent features. 
For example, most of the frequent words and context-insensitive rare words will have vectors with small norms.

\begin{table}[t]
\centering
\begin{tabular}{|l||l|l|}
\hline
Batch  & Len. & Source sentence \\ \hline\hline
\multicolumn{3}{|c|}{\textit{Vanilla}} \\ \hline\hline
\multirow{2}{*}{$\mathcal{B}_1$}  & 16 & In catalogues, magazines $\dots$ \\ 
& 27 & Nevertheless, it is an  $\dots$ \\ \hline
\multirow{2}{*}{$\mathcal{B}_2$}  & 38 & The company ROBERT  $\dots$ \\ 
 & 37 & Ottmar Hitzfeld played  $\dots$ \\ \hline\hline
 \multicolumn{3}{|c|}{\textit{The Proposed Method}} \\ \hline\hline
\multirow{2}{*}{$\mathcal{B}_1^*$}  & 3 & Second Part. \\ 
  & 4 & It was not. \\ \hline
\multirow{2}{*}{$\mathcal{B}_2^*$}   & 5 & Thank you very much. \\ 
  & 4 & We know that. \\ \hline
\end{tabular}
\caption{Training batches on the WMT'14 English--German translation task. ``Len.'' denotes the length of the sentence. The proposed method provides a much easier curriculum at the beginning of the training of the model.}
\label{tab:batchcompair}
\end{table}

Unlike existing CL methods for NMT, relying on a hand-crafted curriculum arrangement~\cite{Zhang:2018vf} or a task-dependent hyperparameter~\cite{cbcl}, the proposed norm-based model competence enables the model to arrange the curriculum itself according to its ability, which is beneficial to practical NMT systems.
We also introduce a novel paradigm to assign levels of difficulty to sentences, as sentence weights, into the objective function for better arrangements of the curricula, enhancing both existing CL systems and the proposed method. 

Empirical results for the two widely-used benchmarks show that the proposed method provides a significant performance boost over strong baselines, while also significantly speeding up the training. 
The proposed method requires slightly changing the data sampling pipeline and the objective function without modifying the overall architecture of NMT, thus no extra parameters are employed.

\section{Background}
NMT uses a single large neural network to construct a translation model that translates a source sentence $\mathbf{x}$ into a target sentence $\mathbf{y}$. 
During training, given a parallel corpus $\mathcal{D}=\{\langle \mathbf{x}^{n},\mathbf{y}^{n}\rangle \}^{N}_{n=1}$, NMT aims to maximize its log-likelihood:
\begin{eqnarray}
    \label{eq:nmtlossfunction}
    \hat{\bm{\theta}} &=& L(\mathcal{D};\bm{\theta}_0) \nonumber \\ 
    &=&\mathop{\arg\max}_{\bm{\theta}_0} \sum_{n=1}^{N}\log P(\mathbf{y}^{n}|\mathbf{x}^{n};\bm{\theta}_0)
\end{eqnarray}
\noindent where $\bm{\theta}_0$ are the parameters to be optimized during the training of the NMT models. Due to the intractability of $N$, the training of NMT employs \textit{mini-batch gradient descent} rather than \textit{batch gradient descent} or \textit{stochastic gradient descent}, as follows:
\begin{eqnarray}
    \label{eq:minibatchvanilla}
    &&\mathcal{B}_1,\cdots,\mathcal{B}_t,\cdots,\mathcal{B}_T=\mathrm{sample}(\mathcal{D}) \\
    &&\hat{\bm{\theta}} = L(\mathcal{B}_T;L(\mathcal{B}_{T-1};\cdots L(\mathcal{B}_1,\bm{\theta}_0)))
\end{eqnarray}
where $T$ denotes the number of training steps and $\mathcal{B}_t$ denotes the $t$th training batch. 
In the training of the $t$th mini-batch, NMT optimizes the parameters $\bm{\theta}_{t-1}$ updated by the previous mini-batch. 

CL supposes that if mini-batches are bucketed in a particular way (e.g. with examples from easy to difficult), this would boost the performance of NMT and speed up the training process as well. That is, upgrading the $\mathrm{sample}(\cdot)$ to
\begin{eqnarray}
    \label{eq:minibatchcl}
    &&\mathcal{B}_1^*,\cdots,\mathcal{B}_t^*,\cdots,\mathcal{B}_T^*=\mathrm{sample}^*(\mathcal{D})
\end{eqnarray}
where the order from easy to difficult (i.e. $\mathcal{B}_1^*\rightarrow\mathcal{B}_T^*$ ) can be: 1) sentences with lengths from short to long; 2) sentences with words whose frequency goes from high to low (i.e. word rarity); and 3) uncertainty of sentences (from low to high uncertainties) measured by models trained in previous epochs or pre-trained language models.
Table~\ref{tab:batchcompair} shows the sentences of the training curricula provided by vanilla Transformer and the proposed method.

\section{Norm-based Curriculum Learning} 
\subsection{Norm-based Sentence Difficulty} 
Most NLP systems have been taking advantage of distributed word embeddings to capture the syntactic and semantic features of a word~\cite{Turian:2010vi,mikolov2013efficient}.
A word embedding (vector) can be divided into two parts: the norm and the direction:
\begin{eqnarray}
    \label{eq:wordvector}
    \bm{w}= \underbrace{||\bm{w}||}_{\mathrm{norm}} \cdot \underbrace{\frac{\bm{w}}{||\bm{w}||}}_{\mathrm{direction}}
\end{eqnarray}
In practice, the word embedding, represented by $\bm{w}$, is the key component of a neural model~\citep{liu2019latent,liu2019shared}, and the direction $\frac{\bm{w}}{||\bm{w}||}$ can also be used to carry out simple word/sentence similarity and relation tasks.
However, the norm $||\bm{w}||$ is rarely considered and explored in the computation.

\begin{figure}[t]
\includegraphics[width=0.48\textwidth]{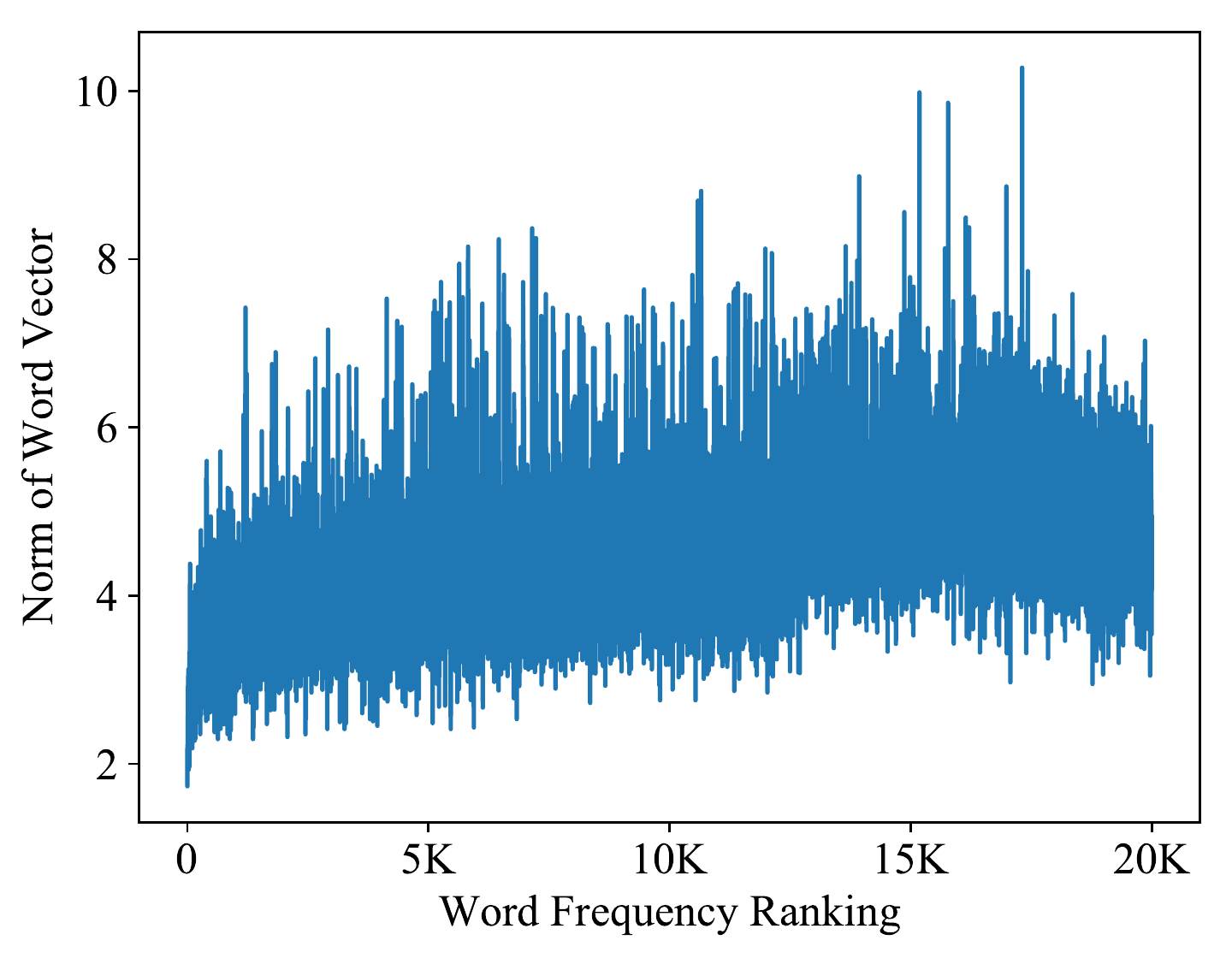}
\caption{Word vector norm of the word embedding model trained on the WMT'14 English--German (source side) training data. The $x$-axis is the word frequency, ranked in descending order. Rare words and significant words have higher norms.}
\label{fig:prewordembeddingnorm}
\end{figure}

Surprisingly, the norm which is simply derived from a single model parameter, can also capture delicate features during the optimization of a model.
\citet{Schakel:2015ub} observe that in the word embedding model~\cite{mikolov2013efficient}, the word vector norm increases with a decrease of the word frequency, while polysemous words, such as ``May'', tend to have an average norm weighted over its various contexts.
\citet{Wilson:2015vv} further conduct controlled experiments on word vector norm and find that besides the word frequency, the diversities of the context of the word are also a core factor to determine its norm.
The vector of a context-insensitive word is assigned a higher norm.
In other words, if a word is usually found in specific contexts, it should be regarded as a significant word~\cite{luhn1958automatic}.
The word embedding model can exactly assign these significant words higher norms, even if some of them are frequent.
The sentences consisting of significant words share fewer commonalities with other sentences, and thus they can also be regarded as difficult-to-learn examples.

Figure~\ref{fig:prewordembeddingnorm} shows the relationship between the word vector norm and the word frequency in the English data of the WMT'14 English--German translation task.
The results stay consistent with prior works~\cite{Wilson:2015vv}, showing that the rare words and significant words obtain a high norm from the word embedding model.
Motivated by these works and our preliminary experimental results, we propose to use the word vector norm as a criterion to determine the difficulty of a sentence.
Specifically, we first train a simple word embedding model on the training corpus, and then obtain an embedding matrix $\mathbf{E}^{\mathrm{w2v}}$. 
Given a source sentence $\mathbf{x}=x_1,\cdots,x_i,\cdots,x_I$, it can be mapped into distributed representations $\bm{x}_1,\cdots,\bm{x}_i,\cdots,\bm{x}_I$ through $\mathbf{E}^{\mathrm{w2v}}$.
The \textbf{norm-based sentence difficulty} is calculated as
\begin{eqnarray}
    \label{eq:sendiff}
    d(\mathbf{x})= \sum_{i=1}^I||\bm{x}_i||
\end{eqnarray}
Long sentences and sentences consisting of rare words or significant words tend to have a high sentence difficulty for CL.

The proposed norm-based difficulty criterion has the following advantages: 1) It is easy to compute since the training of a simple word embedding model just need a little time and CPU resources; 2) Linguistically motivated features, such as word frequency and sentence length, can be effectively modeled; 3) Model-based features, such as learning-dependent word significance, can also be efficiently captured.

\subsection{Norm-based Model Competence}
Besides finding an optimal sentence difficulty criterion, arranging the curriculum in a reasonable order is equally important.
As summarized by~\citet{Zhang:2019wg}, there are two kinds of CL strategies: deterministic and probabilistic.
From their observations, probabilistic strategies are superior to deterministic ones in the field of NMT, benefiting from the randomization during mini-batch training.

Without loss of generality, we evaluate our proposed norm-based sentence difficulty with a typical probabilistic CL framework, that is, competence-based CL~\cite{cbcl}.
In this framework, a notion of model competence is defined which is a function that takes the training step $t$ as input and outputs a competence value from 0 to 1:\footnote{We introduce the square root competence model since it has the best performance in \citet{cbcl}.}
\begin{eqnarray}
    \label{eq:clcomp}
    c(t)\in(0,1]=\min(1, \sqrt{t\frac{1-c_0^2}{\lambda_t} + c_0^2})
\end{eqnarray}
where $c_0=0.01$ is the initial competence at the beginning of training and $\lambda_t$ is a hyperparameter determining the length of the curriculum.
For the sentence difficulty, they use cumulative density function (CDF) to transfer the distribution of sentence difficulties into $(0,1]$:
\begin{eqnarray}
    \label{eq:cdf}
    \hat{d}(\mathbf{x}^n)\in(0,1] = \mathrm{CDF}(\{d(\mathbf{x}^n)\}^N_{n=1})^n
\end{eqnarray}
The score of difficult sentences tends to be 1, while that of easy sentences tends to be 0.
The model uniformly samples curricula whose difficulty is lower than the model competence at each training step, thus making the model learn the curriculum in a probabilistic way.

\begin{figure}[t]
\includegraphics[width=0.5\textwidth]{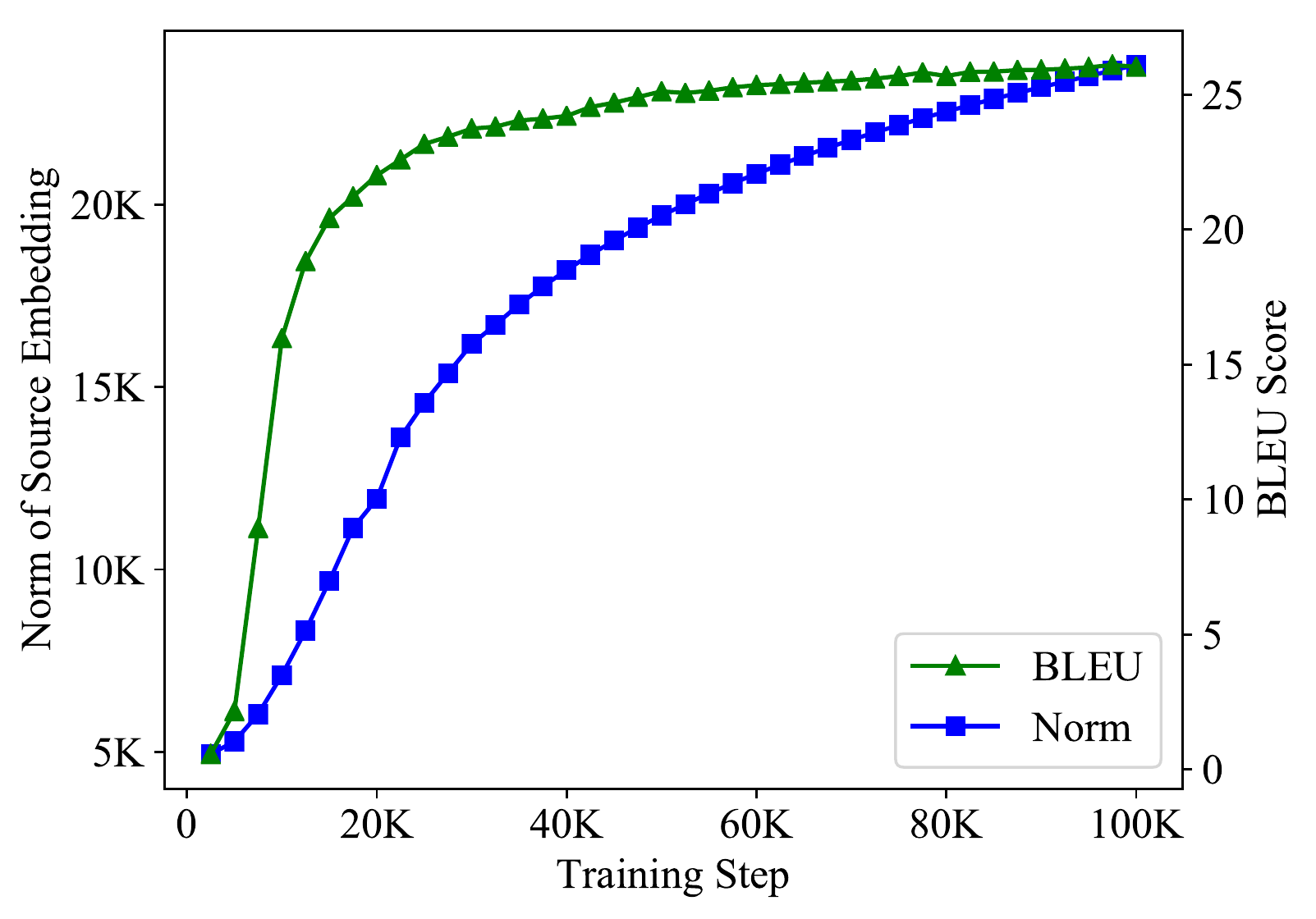}
\caption{Norm of NMT source embedding and BLEU score of a vanilla Transformer on the WMT'14 English--German translation task. The BLEU scores are calculated on the development set. Both the norm and BLEU score grow rapidly until the 30K training step.}
\label{fig:normandbleu}
\end{figure}

One limitation of competence-based CL is that the hyperparameter $\lambda_t$ is task-dependent.
In detail, for each system, it needs to first train a vanilla baseline model and then use the step reaching $90\%$ of its final performance (BLEU score) as the value of the length hyperparameter.
As we know, training an NMT baseline is costly, and arbitrarily initializing the value might lead to an unstable training process.

To alleviate this limitation and enable NMT to learn curricula automatically without human interference in setting the hyperparameter, it is necessary to find a way for the model to determine the length of a curriculum by itself, according to its competence, which should be independent of the specific task.

To this aim, we further introduce a \textbf{norm-based model competence} criterion.
Different from the norm-based difficulty using the word vector norm, the norm-based model competence uses the norm of the source embedding of the NMT model $\mathbf{E}^{\mathrm{nmt}}$:
\begin{eqnarray}
    \label{eq:embnorm}
    m_t= ||\mathbf{E}_t^{\mathrm{nmt}}||
\end{eqnarray}
where $m_t$ denotes the norm of $\mathbf{E}^{\mathrm{nmt}}$ at the $t$th training step, and we write $m_0$ for the initial value of the norm of $\mathbf{E}^{\mathrm{nmt}}$.
This proposal is motivated by the empirical results shown in Figure~\ref{fig:normandbleu}, where we show the BLEU scores and the norms of the source embedding matrix at each checkpoint of a vanilla Transformer model on the WMT'14 English--German translation task.
We found the trend of the growth of the norm $m_t$ to be very similar to that of the BLEU scores.
When $m_t$ stays between 15K to 20K, which is about from twice to three times larger than the initial norm $m_0$, both the growth of the norm and that of the BLEU score have slowed down.
It shows strong clues that $m_t$ is a functional metric to evaluate the competence of the model, and thus we can avoid the intractability of $\lambda_t$ in Equation~\ref{eq:clcomp}:
\begin{eqnarray}
    \label{eq:normcomp}
    \hat{c}(t) =\min(1, \sqrt{(m_t-m_0)\frac{1-c_0^2}{\lambda_m m_0} + c_0^2})
\end{eqnarray}
where $\lambda_m$ is a task-independent hyperparameter to control the length of the curriculum. 
With this criterion, the models can, by themselves, fully automatically design a curriculum based on the feature (norm).
At the beginning of the training, there is a lower $m_t$, so the models tend to learn with an easy curriculum. 
But with an increase of the norm $m_t$, more difficult curricula will be continually added into the learning.

\subsection{Norm-based Sentence Weight}
\label{sec:sentweight}
In competence-based CL, the model uniformly samples sentences whose difficulty level is under the model competence, and then learns with the samples equally.
As a result, those simple sentences with low difficulty (e.g. $\hat{d}(\mathbf{x})<0.1$) are likely to be repeatedly used in the model learning. 
This is somewhat counterintuitive and a waste of computational resources.
For example, when students are able to learn linear algebra, they no longer need to review simple addition and subtraction, but can keep the competence during the learning of hard courses.
On the other hand, a difficult (long) sentence is usually made up of several easy (short) sentences.
Thus, the representations of easy sentences can also benefit from the learning of difficult sentences.

\begin{algorithm*}[t]
  \caption{Norm-based Curriculum Learning Strategy}
  \label{alg:appoarch}
  \begin{algorithmic}[1]
    \Require Parallel corpus $\mathcal{D}=\{\langle \mathbf{x}^{n},\mathbf{y}^{n}\rangle \}^{N}_{n=1}$; Translation system $\theta$;
    \State Train the word2vec Embedding $\mathbf{E}^{\mathrm{w2v}}$ on $\{\mathbf{x}^{n}\}^{N}_{n=1}$.
    \State Compute norm-based sentence difficulty $\{\hat{d}(\mathbf{x}^n)\}^N_{n=1}$ using $\mathbf{E}^{\mathrm{w2v}}$, Eq.~\ref{eq:sendiff} and~\ref{eq:cdf}.
            \For{$t=1$ to $T$} 
        \State Compute norm-based model competence $\hat{c}(t)$ using Eq.~\ref{eq:embnorm} and~\ref{eq:normcomp}.
        \State Generate training batch $\mathcal{B}_t^*$ uniformly sampled from $\{\langle \mathbf{x},\mathbf{y}\rangle|\hat{d}(\mathbf{x})<\hat{c}(t),\langle \mathbf{x},\mathbf{y} \rangle \in \mathcal{D}\}$.
        \State Compute norm-based length weight $\mathcal{W}=\{w(\mathbf{x},t)|\langle \mathbf{x},\mathbf{y} \rangle \in \mathcal{B}_t^*$\} using Eq.~\ref{eq:lengthweight}.
        \State Update $\theta$ with batch loss $\mathbb{E}_{\langle \mathbf{x},\mathbf{y}\rangle \sim  \mathcal{B}_t^*}$ calculated by $\mathcal{W}$ and Eq.~\ref{eq:lossfunc}.
    \EndFor
    \State \Return $\theta$
  \end{algorithmic}
\end{algorithm*}

To alleviate this limitation of competence-based CL and further enhance the learning from the curriculum of different levels of difficulty, we propose a simple yet effective \textbf{norm-based sentence weight}:
\begin{eqnarray}
    \label{eq:lengthweight}
    w(\mathbf{x},t)=(\frac{\hat{d}(\mathbf{x})}{\hat{c}(t)})^{\lambda_w}
\end{eqnarray}
where $\lambda_w$ is the scaling hyperparameter smoothing the weight, $\hat{d}(\mathbf{x})$ is the norm-based sentence difficulty, and $\hat{c}(t)$ is the model competence.
For each training step $t$, or each model competence $\hat{c}(t)$, the weight of a training example $w(\mathbf{x},t)$ is included in its objective function:
\begin{eqnarray}
    \label{eq:lossfunc}
    l(\langle \mathbf{x},\mathbf{y}\rangle,t)=-\log P(\mathbf{y}|\mathbf{x})w(\mathbf{x},t) 
\end{eqnarray}
where $l(\langle \mathbf{x},\mathbf{y}\rangle,t)$ is the training loss of an example $\langle \mathbf{x},\mathbf{y}\rangle$ at the $t$th training step.
With the use of sentence weights, the models, at each training step, tend to learn more from those curricula whose difficulty is close to the current model competence.
Moreover, the models still benefit from the randomization of the mini-batches since the length weight does not change the curriculum sampling pipeline.

\subsection{Overall Learning Strategy}
Algorithm~\ref{alg:appoarch} illustrates the overall training flow of the proposed method.
Besides the component and training flow of vanilla NMT models, only some low-cost operations, such as matrix multiplication, have been included in the data sampling and objective function, allowing an easy implementation as a practical NMT system.
We have also found, empirically, that the training speed of each step is not influenced by the introduction of the proposed method.

\section{Experiments}
\label{sec:exper}
\subsection{Data and Setup}
We conducted experiments on the widely used benchmarks, i.e.~the medium-scale WMT'14 English--German (En-De) and the large-scale WMT'17 Chinese--English (Zh-En) translation tasks.
For En-De, the training set consists of 4.5M sentence pairs with 107M English words and 113M German words.
The development is newstest13 and the test set is newstest14.
For the Zh-En, the training set contains roughly 20M sentence pairs. 
The development is newsdev2017 and the test set is newstest2017.
The Chinese data were segmented by \texttt{jieba},\footnote{\url{https://github.com/fxsjy/jieba}} while the others were tokenized by the \texttt{tokenize.perl} script from Moses.\footnote{\url{http://www.statmt.org/moses/}}
We filtered the sentence pairs with a source or target length over 200 tokens.
Rare words in each data set were split into sub-word units~\citep{DBLP:journals/corr/SennrichHB15}. 
The BPE models were trained on each language separately with 32K merge operations.

All of the compared and implemented systems are the \emph{base} Transformer \cite{DBLP:journals/corr/VaswaniSPUJGKP17} using the open-source toolkit \texttt{Marian}~\cite{mariannmt}.\footnote{\url{https://marian-nmt.github.io/}}
We tie the target input embedding and target output embedding~\cite{Press:2017ug}.
The Adam \cite{kingma2014adam} optimizer has been used to update the model parameters with hyperparameters $\beta_{1}$= 0.9, $\beta_{2}$ = 0.98, $\varepsilon$ = $10^{-9}$.
We use the variable learning rate proposed by \citet{DBLP:journals/corr/VaswaniSPUJGKP17} with 16K warm up steps and a peak learning rate $0.0003$.

\begin{table*}[t]
\centering
\scalebox{1.0}{
\begin{tabular}{|l||l|l|l|l|l|}
\hline
ID & Model & Dev. & Test &Updates &Speedup\\ \hline\hline
\multicolumn{6}{|c|}{\textit{Existing Baselines}} \\ \hline\hline
1 & GNMT~\cite{wu2016google} & - & 24.61 &- & - \\ 
2 & ConvS2S~\cite{DBLP:journals/corr/GehringAGYD17} & - & 25.16 &- & - \\ 
3 & Base Transformer~\cite{DBLP:journals/corr/VaswaniSPUJGKP17} & 25.80 & 27.30 &-&- \\ 
4 & Big Transformer~\cite{DBLP:journals/corr/VaswaniSPUJGKP17}& 26.40 & 28.40  &-&- \\  \hline\hline
\multicolumn{6}{|c|}{\textit{Our Implemented Baselines}} \\ \hline\hline
5 & Base Transformer~\cite{DBLP:journals/corr/VaswaniSPUJGKP17} & 25.90 & 27.64 &100.0K &1.00x \\ 
6 & 5 + Competence-based CL~\cite{cbcl} & 26.39 & 28.19  &60.0K& 1.67x \\  \hline\hline
\multicolumn{6}{|c|}{\textit{Our Proposed Method (Individual)}} \\ \hline \hline
7 & 6 + Norm-based Model Competence  & 26.59 & 28.51 & 50.0K & 2.00x  \\
8 & 6 + Norm-based Sentence Complexity  & 26.61 & 28.61 & 50.0K & 2.00x \\
9 & 6 + Norm-based Sentence Weight  & 26.63 & 28.32 & 52.5K & 1.90x  \\  \hline \hline
\multicolumn{6}{|c|}{\textit{Our Proposed Method (All)}} \\ \hline \hline
10 & 5 + Norm-based CL   &   \textbf{26.89} & \textbf{28.81}  & \textbf{45.0K} & \textbf{2.22x}  \\ \hline
\end{tabular}}
\caption{Results on the WMT'14 English--German translation task. Dev. is the newstest2013 while Test is newstest2014. `Updates' means the step of each model reaching the best performance of model (5) (K = thousand), while `Speedup' means its corresponding speedup.}

\label{tab:mainresults}
\end{table*}
\begin{table}[t]
\centering
\scalebox{1.0}{
\begin{tabular}{|l|l||l|l|}
\hline
$\lambda_m$ & Dev. & $\lambda_w$ & Dev. \\ \hline\hline
1.0 & 26.63 & 0   & 26.71 \\ \hline
2.0 & 26.72 & 1/3 & 26.80 \\ \hline
2.5 & \textbf{26.89} & 1/2 & \textbf{26.89} \\ \hline
3.0 & 26.65 & 1   & 26.78 \\   \hline
4.0 & 26.62 & 2   & 26.77 \\   \hline
\end{tabular}}
\caption{Effects of different $\lambda_m$ of the norm-based model competence function and $\lambda_w$ of the norm-based sentence weight function.}
\label{tab:abaltion}
\end{table}

We employed \texttt{FastText}~\cite{Bojanowski:2017un}\footnote{\url{https://github.com/facebookresearch/fastText}} with its default settings to train the word embedding model for calculating the norm-based sentence difficulty; an example is given in Figure~\ref{fig:prewordembeddingnorm}.
The hyperparameters $\lambda_m$ and $\lambda_w$ controlling the norm-based model competence and norm-based sentence weight were tuned on the development set of En-De, with the value of 2.5 and 0.5, respectively.
To test the adaptability of these two hyperparameters, we use them directly for the Zh-En translation task without any tuning. 
We compare the proposed methods with the re-implemented competence-based CL~\cite{cbcl}.\footnote{We use its best settings, i.e.~the rarity-based sentence difficulty and the square root competence function.}

During training, the mini-batch contains nearly 32K source tokens and 32K target tokens.
We evaluated the models every 2.5K steps, and chose the best performing model for decoding.
The maximum training step was set to 100K for En-De and 150K for Zh-En.
During testing, we tuned the beam size and length penalty~\cite{wu2016google} on the development data, using a beam size of 6 and a length penalty of 0.6 for En-De, and a beam size of 12 and a length penalty of 1.0 for Zh-En.
We report the 4-gram BLEU \citep{papineni2002bleu} score given by the \textit{multi-bleu.perl} script.
The codes and scripts of the proposed norm-based CL and our re-implemented competence-based CL are freely available at \url{https://github.com/NLP2CT/norm-nmt}.

\subsection{Main Results}
Table~\ref{tab:mainresults} shows the results of the En-De translation task in terms of BLEU scores and training speedup.
Models (1) to (4) are the existing baselines of this translation benchmark.
Model (5) is our implemented base Transformer with 100K training steps, obtaining 27.64 BLEU scores on the test set.
By applying the competence-based CL with its proposed sentence rarity and square root competence function, i.e. model (6), it reaches the performance of model (5) using 60K training steps and also gets a better BLEU score.

For the proposed method, we first show the performance of each sub-module, that is: model (7), which uses the norm-based model competence instead of the square root competence of model (6); model (8), which uses the proposed norm-based sentence complexity instead of the sentence rarity of model (6); and model (9), which adds the norm-based sentence weight to model (6).
The results show that after applying each sub-module individually, both the BLEU scores and the learning efficiency are further enhanced.

\begin{table*}[t]
\centering
\scalebox{1.0}{
\begin{tabular}{|l||l|l|l|l|l|}
\hline
 ID &Model & Dev. & Test &Updates &Speedup \\ \hline\hline
\multicolumn{6}{|c|}{\textit{Existing Baselines}} \\ \hline \hline
11 & Base Transformer~\cite{Ghazvininejad:2019wz} & - & 23.74 & - & -\\ \hline
12 & Big Transformer~\cite{Ghazvininejad:2019wz} & - & 24.65 & -& - \\ \hline\hline
 \multicolumn{6}{|c|}{\textit{Our Implemented Baselines}} \\ \hline \hline
13 & Base Transformer~\cite{DBLP:journals/corr/VaswaniSPUJGKP17} & 22.29 & 23.69 &150.0K  & 1.00x\\ \hline
14 & 13+Competence-based CL~\cite{cbcl} & 22.75 & 24.30 &60.0K  & 2.50x\\ \hline\hline
\multicolumn{6}{|c|}{\textit{Our Proposed Method}} \\ \hline \hline
15 & 13+Norm-based CL & \textbf{23.41} & \textbf{25.25} & \textbf{45.0K}  & \textbf{3.33x}\\ \hline
\end{tabular}}
\caption{Results on the large-scale WMT'17 Chinese--English translation task. Dev. is the newsdev2017 while Test is newstest2017. `Updates' means the step of each model reaching the best performance of model (13) (K = thousand), while `Speedup' means its corresponding speedup.}
\end{table*}

\begin{figure*}[t]%
\centering
    \subfigure[Length]{
    \scalebox{0.52}{\begin{tikzpicture}
\Large
\definecolor{c1}{RGB}{238,59,57}
\definecolor{c2}{RGB}{94,170,95}
\definecolor{c3}{RGB}{254,206,0}
\definecolor{c4}{RGB}{157,106,185}
\begin{axis}[
    ylabel={BLEU Score},
    xmin=-10, xmax=60,
    ymin=26, ymax=30,
    xticklabels={Short,Medium,Long},
    xtick={0,25,50},
    ytick={26,27,28,29,30},
    legend pos=north west,
    ymajorgrids=true,
    grid style=dashed,
    legend cell align={left}
]
\addplot[thick,
    color=blue,
    mark=square*,
    ]
    coordinates {
    (0,27.39)(25,27.88)(50,29.67)
    };
\addplot[thick,
    color=red,
    mark=triangle*,
    ]
    coordinates {
    (0,27.45)(25,27.56)(50,29.02)
    };
\addplot[thick,
    color=black!40!green,
    mark=oplus*,
    ]
    coordinates {
    (0,27.74)(25,27.12)(50,29.05)
    };
\addplot[thick,
    color=c3,
    mark=diamond*,
    ]
    coordinates {
    (0,26.16)(25,27.07)(50,28.24)
    };

    \legend{NBCL,CBCL+NBSW,CBCL,Vanilla}
\end{axis}
\end{tikzpicture}}
    \label{fig:comparelength}}
\hfill
\subfigure[Frequency]{%
    \scalebox{0.52}{\begin{tikzpicture}
\Large
\definecolor{c1}{RGB}{238,59,57}
\definecolor{c2}{RGB}{94,170,95}
\definecolor{c3}{RGB}{254,206,0}
\definecolor{c4}{RGB}{157,106,185}
\begin{axis}[
    xmin=-10, xmax=60,
    ymin=26, ymax=30,
    xticklabels={Frequent,Medium,Rare},
    xtick={0,25,50},
    ytick={26,27,28,29,30},
    legend pos=north west,
    ymajorgrids=true,
    grid style=dashed,
    legend cell align={left}
]
\addplot[thick,
    color=blue,
    mark=square*,
    ]
    coordinates {
    (0,27.05)(25,28.14)(50,29.65)
    };
\addplot[thick,
    color=red,
    mark=triangle*,
    ]
    coordinates {
    (0,26.94)(25,27.80)(50,29.05)
    };
\addplot[thick,
    color=black!40!green,
    mark=oplus*,
    ]
    coordinates {
    (0,27.28)(25,27.24)(50,29.14)
    };
\addplot[thick,
    color=c3,
    mark=diamond*,
    ]
    coordinates {
    (0,26)(25,27.4)(50,28.14)
    };

\end{axis}
\end{tikzpicture}}
    \label{fig:comparecl}}
\hfill
\subfigure[Norm]{%
    \scalebox{0.52}{\begin{tikzpicture}
\Large
\definecolor{c1}{RGB}{238,59,57}
\definecolor{c2}{RGB}{94,170,95}
\definecolor{c3}{RGB}{254,206,0}
\definecolor{c4}{RGB}{157,106,185}
\begin{axis}[
    xmin=-10, xmax=60,
    ymin=26, ymax=30,
    xticklabels={Simple,Medium,Hard},
    xtick={0,25,50},
    ytick={26,27,28,29,30},
    legend pos=north west,
    ymajorgrids=true,
    grid style=dashed,
    legend cell align={left}
]
\addplot[thick,
    color=blue,
    mark=square*,
    ]
    coordinates {
    (0,26.93)(25,28.36)(50,29.56)
    };
\addplot[thick,
    color=red,
    mark=triangle*,
    ]
    coordinates {
    (0,26.93)(25,27.91)(50,28.99)
    };
\addplot[thick,
    color=black!40!green,
    mark=oplus*,
    ]
    coordinates {
    (0,27.19)(25,27.71)(50,28.9)
    };
\addplot[thick,
    color=c3,
    mark=diamond*,
    ]
    coordinates {
    (0,26)(25,27.61)(50,28.03)
    };

\end{axis}
\end{tikzpicture}}
    \label{fig:comparenorm}}
\caption{Translation performance of each NMT system in (a) length-based, (b) frequency-based, and (c) norm-based difficulty buckets. The reported BLEU scores are evaluated on the three subsets evenly divided by the En-De test set based on sentence difficulty. NBCL and CBCL denote norm-based and competence-based CL, respectively. CBCL+NBSW denotes the integration of norm-based sentence weight and competence-based CL.}
\label{fig:wo}%
\end{figure*}
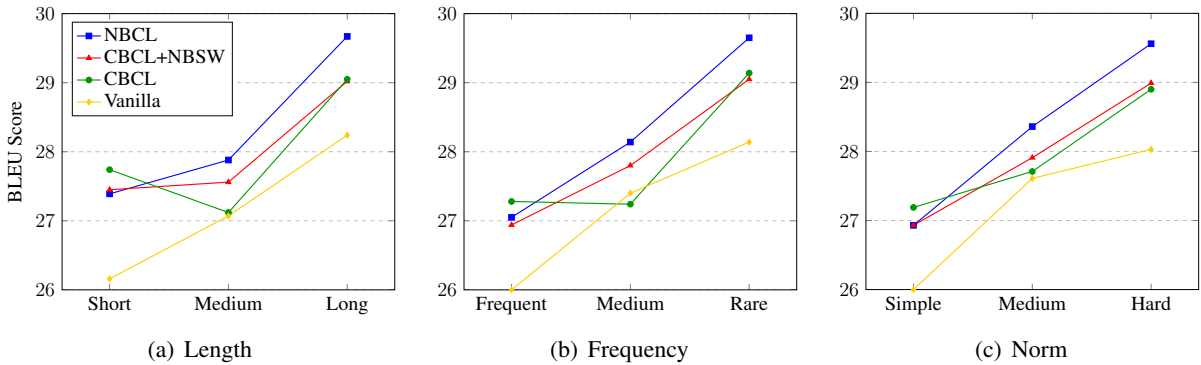

Model (10) shows the results combining the three proposed norm-based methods for CL, i.e.~the norm-based sentence difficulty, model competence, and sentence weight. 
We call the combination of the proposed method norm-based CL.
It shows its superiority in the BLEU score, which has an increase of 1.17 BLEU scores compared to the Transformer baseline, as well as speeding up the training process by a factor of 2.22. 
One can note that all of our implemented systems have the same number of model parameters; besides, the training step of each model involves essentially the same execution time, resulting in a deployment-friendly system.

\subsection{Effect of $\lambda_m$ and $\lambda_w$}
Table~\ref{tab:abaltion} shows the effects of the two hyperparameters used in the proposed method.
For each experiment, we kept the other parameters unchanged and only adjusted the hyperparameter.
For $\lambda_m$, controlling curriculum length, the higher the value, the longer the curriculum length.
When setting $\lambda_m$ to 2.5 with the curriculum length of nearly 29K steps, it achieves the best performance.
For $\lambda_w$, the scaling sentence weight of the objective function, one achieves satisfactory results with a value of 0.5, which maintains the right balance between the learning of simple and hard examples.

\begin{table*}[t]
    \centering
    \scalebox{1.0}{
    \begin{tabular}{|l||p{0.83\textwidth}|}
        \hline
        Source & Last year a team from the University of Lincoln found that dogs turn their heads to the left when looking at an aggressive dog \textbf{and to the right when looking at a happy dog}. \\
        Reference & Letztes Jahr fand ein Team der Universität von Lincoln heraus, dass Hunde den Kopf nach links drehen, wenn sie einen aggressiven Hund ansehen, \textbf{und nach rechts, wenn es sich um einen zufriedenen Hund handelt}. \\ \hline\hline
        Vanilla & Im vergangenen Jahr stellte ein Team der Universität Lincoln fest, dass Hunde beim Blick auf einen aggressiven Hund nach links abbiegen. \\
        NBCL & Letztes Jahr fand ein Team von der Universität von Lincoln heraus, dass Hunde ihren Kopf nach links drehen, wenn sie einen aggressiven Hund sehen  \textbf{und rechts, wenn sie einen glücklichen Hund sehen}.\\
        \hline
    \end{tabular}}
    \caption{Example of a translation which is regarded as a difficult sentence in terms of the norm-based sentence difficulty, from the En-De test set. The vanilla Transformer omits translating the \textbf{bold} part of the source.}
    \label{tab:translationexmaple}    
\end{table*}

\subsection{Results on the Large-scale NMT}
Although the hyperparameters $\lambda_m$ and $\lambda_w$ have been sufficiently validated on the En-De translation, the generalizability of the model trained using these two hyperparameters is still doubtful.
To clear up any doubts, we further conducted the experiments on the large-scale Zh-En translation without tuning these two hyperparameters, that is, directly using $\lambda_m=2.5$ and $\lambda_w=0.5$.
Specifically, the only difference is the use of a large number of training steps in Zh-En, namely, 150K, for the purpose of better model fitting.

We first confirm the effectiveness of competence-based CL in large-scale NMT, that is model (14), which shows both a performance boost and a training speedup.
Model (15), which trains NMT with the proposed norm-based CL, significantly improves the BLEU score to 25.25 (+1.56) and speeds up the training by a factor of 3.33, showing the generalizability of the proposed method.
The results show that large-scale NMT obtains a greater advantage from an orderly curriculum with enhanced representation learning.
The proposed norm-based CL enables better and faster training of large-scale NMT systems.

\subsection{Effect of Sentence Weight}
As discussed in Section~\ref{sec:sentweight}, competence-based CL over-trains on the simple curriculum, which might lead to a bias in the final translation.
To verify this, we quantitatively analysed the translations generated by different systems.
Figure~\ref{fig:wo} presents the performance of the vanilla Transformer, and of the NMTs trained by competence-based CL and norm-based CL. 
By dividing the En-De test set (3,003 sentences) into three subsets (1001 sentences) according to the length-based sentence difficulty, the frequency-based sentence difficulty, and the norm-based sentence difficulty, we calculated the BLEU scores of each system on each subset.

The results confirm our above assumption, although competence-based CL performs much better in translating simple sentences due to its over-training, the translation of sentences of medium difficulty worsens.
However, the norm-based CL benefits from the norm-based sentence weight, successfully alleviating this issue by applying a scale factor to the loss of simple curricula in the objective function, leading to a consistently better translation performance over the vanilla Transformer.

To further prove the effectiveness of the proposed norm-based sentence weight, we explore the model integrating norm-based sentence weight with competence-based CL, and find that it can also strike the right balance between translating simple and medium-difficulty sentences.

\subsection{A Case Study}
Table~\ref{tab:translationexmaple} shows an example of a translation of a difficult sentence consisting of several similar clauses in the norm-based difficulty bucket.
We observe that the translation by the vanilla model omits translating the last clause, but NMT with norm-based CL translates the entire sentence.
The proposed method enhances the representation learning of NMT, leading to better understandings of difficult sentences, thus yielding better translations.

\section{Related Work}
The norm of a word embedding has been sufficiently validated to be highly correlated with word frequency. 
\citet{Schakel:2015ub} and \citet{Wilson:2015vv} train a simple word embedding model~\cite{mikolov2013efficient} on a monolingual corpus, and find that the norm of a word vector is relevant to the frequency of the word and its context sensitivity: frequent words and words that are insensitive to context will have word vectors of low norm values.

For language generation tasks, especially NMT, there is still a correlation between word embedding and word frequency.
\citet{Gong:2018ul} observe that the word embedding of NMT contains too much frequency information, considering two frequent and rare words that have a similar lexical meaning to be far from each other in terms of vector distance.
\citet{gao2018representation} regard this issue as a representation degeneration issue that it is hard to learn expressive representations of rare words due to the bias in the objective function.
\citet{nguyen-chiang-2018-improving} observe a similar issue during NMT decoding: given two word candidates with similar lexical meanings, NMT chooses the more frequent one as the final translation.
They attribute this to the norm of word vector, and find that target words with different frequencies have different norms, which affects the NMT score function.
In the present paper, for the sake of obtaining an easy and simple word vector norm requirement, we use the norm derived from a simple word embedding model.
In the future, we would like to test norms of various sorts.

There are two main avenues for future research regarding CL for NMT: sentence difficulty criteria and curriculum training strategies.
Regarding sentence difficulty, there are linguistically motivated features~\citep{Kocmi:2017tw,cbcl} and model-based features~\citep{Zhang:2016ue,Kumar:2019vn,Zhang:2019wg,UACL20}.
Both types of difficulty criteria have their pros and cons, while the proposed norm-based sentence difficulty takes the best of both worlds by considering simplicity and effectiveness at the same time.

Regarding the training strategy, both deterministic~\cite{Zhang:2016ue,Kocmi:2017tw} and probabilistic strategies~\cite{cbcl,Zhang:2019wg,Kumar:2019vn} can be better than the other, depending on the specific scenario.
The former is easier to control and explain, while the latter enables NMT to benefit from the randomization of mini-batch training.
However, both kinds of strategy need to carefully tune the CL-related hyperparameters, thus making the training process somewhat costly.
In the present paper, we have designed a fully automated training strategy for NMT with the help of vector norms, removing the need for manual setting.

\section{Conclusion}
We have proposed a novel norm-based curriculum learning method for NMT by: 1) a novel sentence difficulty criterion, consisting of linguistically motivated features and learning-dependent features; 2) a novel model competence criterion enabling a fully automatic learning framework without the need for a task-dependent setting of a feature; and 3) a novel sentence weight, alleviating any bias in the objective function and further improving the representation learning.
Empirical results on the medium- and large-scale benchmarks confirm the generalizability and usability of the proposed method, which provides a significant performance boost and training speedup for NMT.

\section*{Acknowledgements}
This work was supported in part by the National Natural Science Foundation of China (Grant No. 61672555), the Joint Project of the Science and Technology Development Fund, Macau SAR and National Natural Science Foundation of China (Grant No. 045/2017/AFJ), the Science and Technology Development Fund, Macau SAR (Grant No. 0101/2019/A2), and the Multi-year Research Grant from the University of Macau (Grant No. MYRG2017-00087-FST). We thank the anonymous reviewers for their insightful comments. 

\bibliography{acl2020}
\bibliographystyle{acl_natbib}

\end{document}